\documentclass[10pt,twocolumn,letterpaper]{article}
\usepackage{cvpr}
\usepackage{times}

\usepackage{epsfig}
\usepackage{graphicx}
\usepackage{amsmath}
\usepackage{amssymb}

\usepackage[dvipsnames]{xcolor}
\usepackage{float}
\usepackage{adjustbox}
\usepackage{soul}
\usepackage{subcaption}
\usepackage{pifont}
\usepackage{float}

\usepackage{multirow}
\graphicspath{ {figures} }
\usepackage[pagebackref=true,breaklinks=true,letterpaper=true,colorlinks,bookmarks=false]{hyperref}
\definecolor{mypink1}{rgb}{0.858, 0.188, 0.478}

\definecolor{olivegreen}{rgb}{0.33, 0.42, 0.18}

\newcommand{\myparagraph}[1]{\vspace{6pt}\noindent{\bf #1}}

\definecolor{armygreen}{rgb}{0.29, 0.33, 0.13}

\cvprfinalcopy 

\def\cvprPaperID{3746} 

\begin{document}

\title{Generalized Zero- and Few-Shot Learning via Aligned Variational Autoencoders}

\author{
 Edgar Sch{\"o}nfeld$^{1}$ \hspace{4mm} Sayna Ebrahimi$^{2}$ \hspace{4mm} Samarth Sinha$^{3}$ \hspace{4mm} Trevor Darrell$^{2}$ \hspace{4mm} Zeynep Akata$^{4}$\vspace{4mm} \\ 
  \begin{tabular}{cccc}
  $^{1}$Bosch Center for AI & $^{2}$UC Berkeley  &  $^{3}$ University of Toronto & $^{4}$University of Amsterdam 
 \end{tabular}
 }

\maketitle

\begin{abstract}
    %
    Many approaches in generalized zero-shot learning rely on cross-modal mapping between the image feature space and the class embedding space. As labeled images are expensive, one direction is to augment the dataset by generating either images or image features. However, the former misses fine-grained details and the latter requires learning a mapping associated with class embeddings. In this work, we take feature generation one step further and propose a model where a shared latent space of image features and class embeddings is learned by modality-specific aligned variational autoencoders. This leaves us with the required discriminative information about the image and classes in the latent features, on which we train a softmax classifier.
  The key to our approach is that we align the distributions learned from images and from side-information to construct latent features that contain the essential multi-modal information associated with unseen classes. We evaluate our learned latent features on several benchmark datasets, i.e. CUB, SUN, AWA1 and AWA2, and establish a new state of the art on generalized zero-shot as well as on few-shot learning. Moreover, our results on ImageNet with various zero-shot splits show that our latent features generalize well in large-scale settings.
\end{abstract}

\section{Introduction}
Generalized zero-shot learning (GZSL) is a challenging task especially for unbalanced and large datasets such as ImageNet~\cite{deng2009imagenet}. Although at training time no visual data of some classes, i.e. unseen classes, are provided the classifier must learn to differentiate between all classes, i.e. seen and unseen classes. As visual data of unseen classes is not available at training time, typically knowledge transfer from seen to unseen classes is achieved via some form of side information that encode semantic relationship between classes, i.e. class embeddings. 

\begin{figure}[t]
\centering
\includegraphics[width=1.0\linewidth]{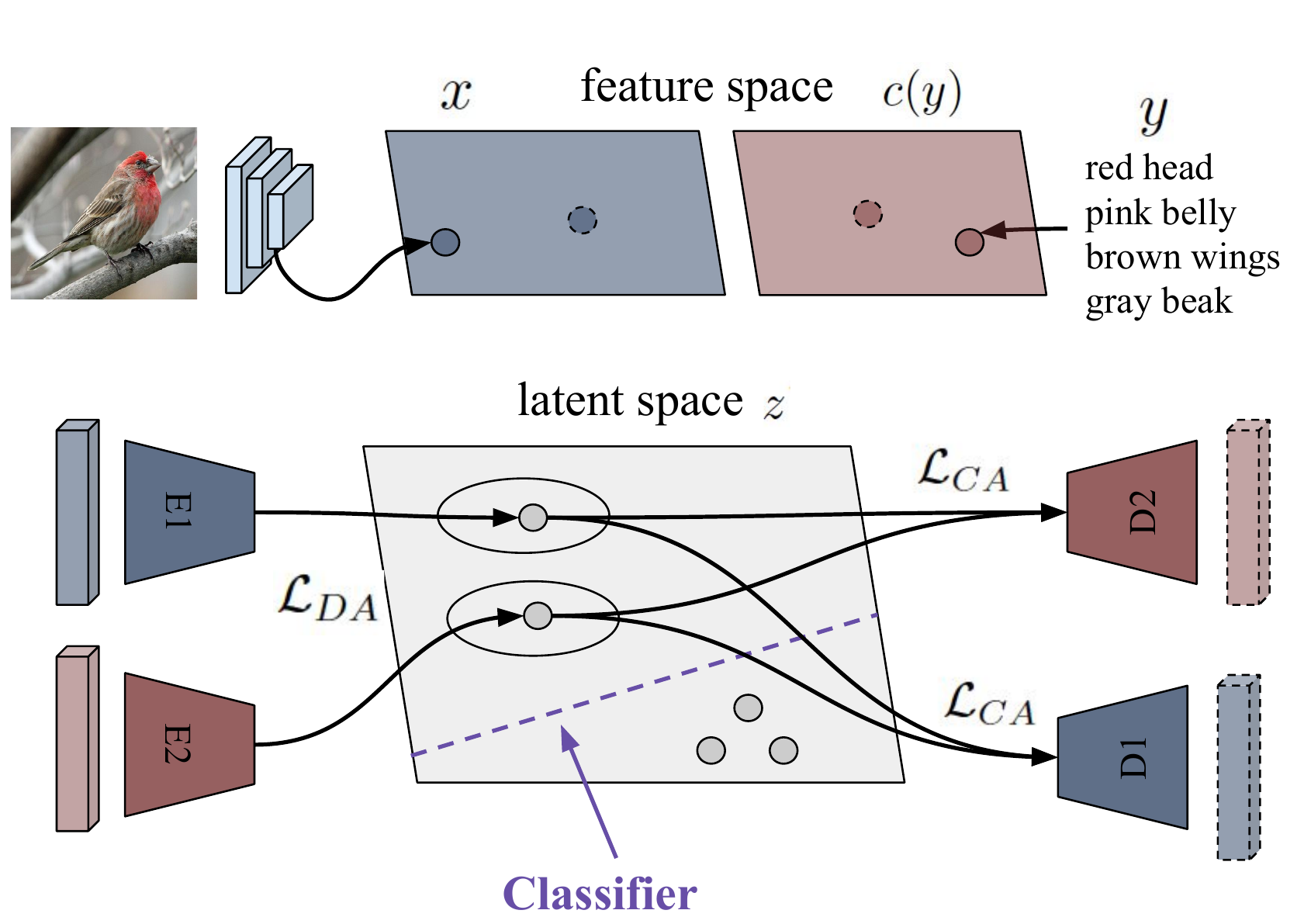}
\caption{Our \texttt{CADA-VAE} model learns a latent embedding ($z$) of image features ($x$) and class embedding ($c(y)$ of labels $y$) via aligned VAEs optimized with cross-alignment ($\mathcal{L}_{CA}$) and distribution alignment ($\mathcal{L}_{DA}$) objectives, and subsequently trains a classifier on sampled latent features of seen and unseen classes. }
\label{fig:teaser}
\end{figure}

Most approaches to GZSL~\cite{devise,ss,norouzi2013zero,latem,sje} learn a mapping between images and their class embeddings.
An orthogonal approach is to augment data by generating artificial images~\cite{reed}. However, due to the level of detail missing in the synthetic images, CNN features extracted from them do not improve classification accuracy. To alleviate this issue, \cite{featgen} proposed to generate image features via a conditional WGAN, which simplifies the task of the generative model and directly optimizes the loss on image features. Although the features generated by~\cite{featgen} improved GZSL significantly, GAN-based loss functions suffer from instability in training. Hence, recently conditional variational autoencoders (VAE)~\cite{cvae,segzsl} have been employed for this purpose. As GZSL is inherently a multi-modal learning task, \cite{tsai2017learning} proposed to transform both modalities to the latent spaces of autoencoders and match the corresponding distributions by minimizing the Maximum Mean Discrepancy (MMD).
Learning such cross-modal embeddings is beneficial for potential downstream tasks that require multimodal fusion, e.g. visual question answering. In this domain, \cite{ramakrishnan2017empirical} recently used a cross-modal autoencoder to extend visual question answering to previously unseen objects.

In this work, we train VAEs to encode and decode features from different modalities, e.g. images and class attributes, and use the learned latent features to train a generalized zero-shot learning classifier. Our latent representations are aligned by matching their parametrized distributions and by enforcing a cross-modal reconstruction criterion.
Consequently, by explicitly enforcing alignment both in the latent features and in the distributions of latent features learned using different modalities, the VAEs enable knowledge transfer to unseen classes without forgetting the previously seen classes.

Our contributions are as follows. (1) We propose the CADA-VAE model that learns shared cross-modal latent representations of multiple data modalities using VAEs via distribution alignment and cross alignment objectives. (2) We extensively evaluate our model using conventional benchmark datasets, i.e. CUB, SUN, AWA1 and AWA2, on zero-shot and few-shot learning settings. Our model establishes the new state-of-the-art performance on generalized zero-shot and few-shot learning settings on all these datasets. Furthermore, we show that our model can be extended easily to more than two modalities that are trained simultaneously. (3) Finally, we show that the latent features learned by our model improve the state of the art in the truly large-scale ImageNet dataset in all splits for the generalized zero-shot learning task.

\section{Related Work}
\label{sec:rel}
In this section, we present related work on generalized zero-shot learning, few-shot learning and cross-modal reconstruction.

\myparagraph{Generalized Zero-and Few-Shot Learning.} In zero-shot learning, training and test classes are disjoint with shared attributes annotated on class level, and the performance of the method is solely judged on its classification accuracy on the novel, i.e. unseen classes. Generalized zero-shot learning is a more realistic variant of zero-shot learning, since the same information is available at training time, but the performance of the model is judged on the harmonic mean of the classification accuracy on seen and unseen classes.

In few-shot learning, there are $k$ examples provided at training time for the previously unseen classes \cite{tsai2017improving, snell2017prototypical, hallucinating, wang2018low}. Using auxiliary information for few-shot learning was introduced in \cite{tsai2017improving}, where attributes related to images were used to improve the performance of the model. The use of auxiliary information was also explored in ReViSE~\cite{tsai2017learning}, in which a common image-label semantic space for transductive few-shot learning is learned. Analogous to the relation between ZSL and GZSL, we extend few-shot to the generalized few-shot learning (GFSL) setting, in which we evaluate the model on both seen and unseen classes.

\myparagraph{Data-Generating Models for GZSL.} Generative models are used for generating images or image feature as a data-augmentation~\cite{featgen,cvae,segzsl} mechanism in GZSL. These approaches treat GZSL as a missing data problem and train conditional GANs or conditional VAEs to generate image features for unseen classes from semantic side-information. In this work, we create latent space features instead.

\myparagraph{Cross-Modal Embedding Models.}
Recent cross-modal embeddings for GZSL are based on autoencoders, such as ReViSE~\cite{tsai2017learning} and DMAE~\cite{DMAE}, which learn to jointly represent features from different modalities in their latent space. By making use of autoencoders, it is possible to learn representations of visual and semantic information in a semi-supervised fashion.
Learning a joint representation for visual and semantic data is achieved by aligning the latent distributions between different data types. ReViSE implements this distribution alignment by minimizing the maximum mean discrepancy between the two latent distributions~\cite{gretton2012kernel}. DMAE aligns distributions by means of minimizing the Squared-Loss Mutual Information~\cite{smi}.
In this work, we use Variational Autoencoders instead, and align the latent distributions by minimizing their Wasserstein distance. In contrast to \cite{DMAE} and \cite{tsai2017learning}, we also enforce a cross-reconstruction loss, by decoding every encoded feature into every other modality.

\begin{figure*}[t]
\centering
\includegraphics[width=0.9\linewidth]{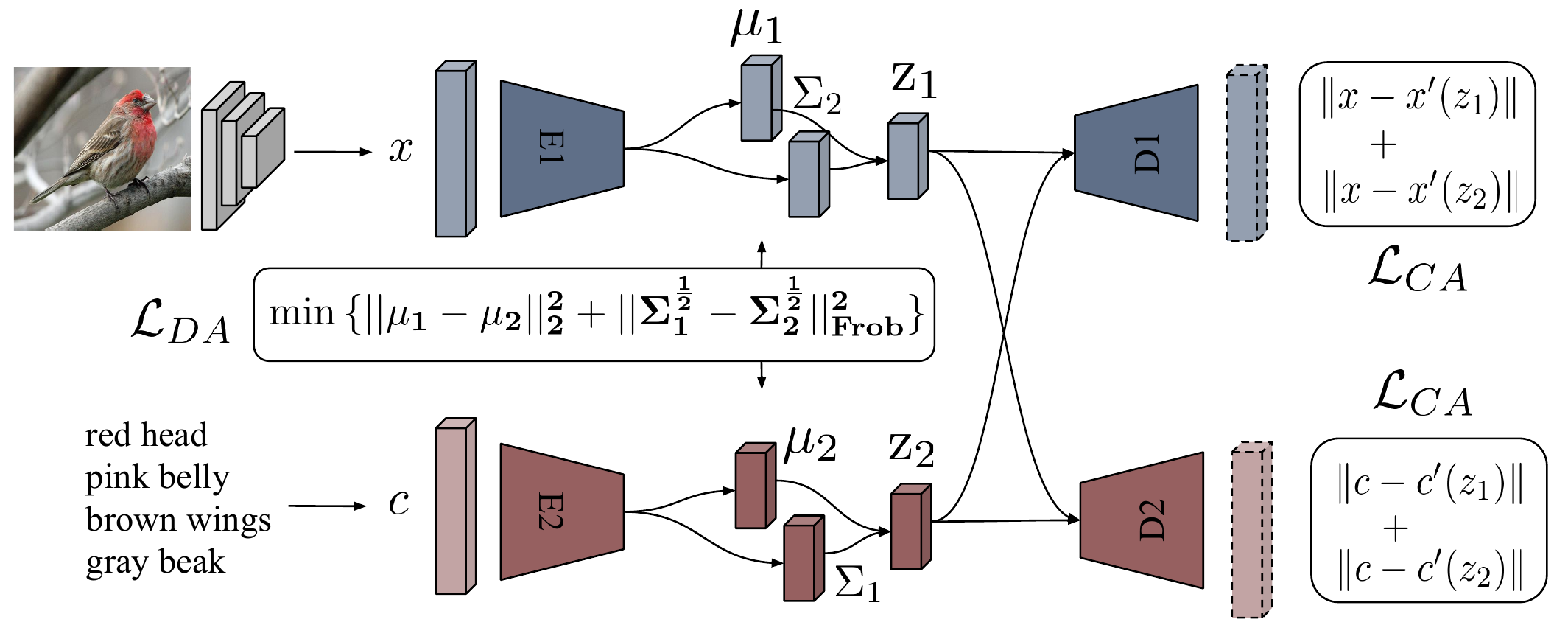}
\caption{Our Cross- and Distribution Aligned VAE (\texttt{CADA-VAE} ). Latent distribution alignment is achieved by minimizing the Wasserstein distance between the latent distributions ($\mathcal{L}_{DA}$). Similarly, the cross-alignment loss ($\mathcal{L}_{CA}$) encourages the latent distributions to align through cross-modal reconstruction.} 
\label{fig:crossvae}
\end{figure*}

\myparagraph{Cross-Reconstruction in Generative models.}
Reconstructing data across domains, referred to as \textit{cross-alignment}, is commonly used in the field of domain adaptation. While models like CycleGAN \cite{zhu2017unpaired}  learn to generate data across domains directly, latent space models use cross-reconstruction to capture the common information contained in both domains in their intermediate latent representations \cite{adda}. In this regard, cross-aligned VAE's have been used previously for text-style transfer~\cite{styletransfer} and image-to-image translation~\cite{imagetranslation}. In \cite{styletransfer} a cross-aligned VAE ensures that the latent representations of texts from different input domains are similar, while in \cite{imagetranslation} a comparable approach matches the latent representations of images from different domains. Both methods have in common that they use a different variant of VAEs with an adversarial loss. Additionally, \cite{styletransfer} makes use of conditional encoders and decoders, while \cite{imagetranslation} enforces cycle consistency and weight sharing. Similarly, better generalization can be achieved if the shared representation space is amenable to class-interpolation, sentence interpolation~\cite{bowman2016generating} and image interpolation~\cite{higgins2016beta}.
In this paper, our building blocks are unconditional VAEs and we achieve multi-modal alignment via cross-reconstruction and latent distribution alignment in a highly reduced space.

\section{CADA-VAE Model}
\label{sec:model}
Of the existing GZSL models, recent data generating approaches~\cite{featgen,segzsl,cvae} achieve superior performance over other methods on disjoint datasets.
Classifying generated image features from GANS~\cite{featgen} or conditional VAEs \cite{segzsl,cvae} runs at risk of being compromised by the curse of dimensionality. On the other hand, \texttt{CADA-VAE} has a control over the dimensionality and structure (via a prior) of the features to be classified.
The main insight of our proposed model is that instead of generating images or image features, we generate low-dimensional latent features and achieve both stable training and state-of-the-art performance. Hence, the key to our approach is the choice of a VAE latent-space, a reconstruction and cross-reconstruction criterion to preserve class-discriminative information in lower dimensions, as well as explicit distribution alignment to encourage domain-agnostic representations.

\subsection{Background}

We first provide the background as the task (GZSL) and as the model building blocks (variational autoencoders).

\myparagraph{Generalized Zero-shot Learning.} The task definition of GZSL is as follows.
Let
 $S = \{(x,y,c(y))|\ x \in X,\ y \in Y^S,\ c(y) \in C \} $
be a set of training examples, consisting of image-features $x$, e.g. extracted by a CNN, class labels $y$ available during training and class-embeddings $c(y)$. Typical class-embeddings are vectors of hand-annotated continuous attributes or Word2Vec features \cite{mikolov2013distributed}. In addition, an auxiliary training set
$
U = \{(u,c(u))|\ u \in Y^u,\  c(u) \in C\}
$
is used, where $u$ denote unseen classes from a set $Y^u$, which is disjoint from $Y^S$. Here,  $C(U)=\{c(u_1),...,c(u_L)\}$ is the set of class-embeddings of unseen classes.
In the legacy challenge of ZSL, the task is to learn a classifier $f_{ZSL}: X \rightarrow Y^U$. However, in this work, we focus on the more realistic and challenging setup of GZSL where the aim is to learn a classifier $f_{GZSL}: X \rightarrow Y^U \cup Y^S$.

\myparagraph{Variational Autoencoder (VAE).}
The basic building block of our model is the variational autoencoder (VAE)~\cite{kingmawelling}.
Variational inference aims at finding the true conditional probability distribution over the latent variables, $p_\phi(z|x)$. Due to interactibility of this distribution, it can be approximated by finding its closest proxy posterior, $q_\theta(z|x)$, through minimizing their distance using a variational lower bound limit. The objective function of a VAE is the variational lower bound on the marginal likelihood of a given datapoint and can be formulated as:
\begin{align}
\mathcal{L} = \mathbb{E}_{q_\phi(z|x)}[\log p_\theta(x|z)]
- D_{KL}( q_\phi(z|x)||p_\theta(z) )
\end{align}
where the first term is the reconstruction error and the second term is the unpacked Kullback-Leibler divergence between the inference model $q(z|x)$, and $p(z)$. A common choice for the prior is a multivariate standard Gaussian distribution. The encoder predicts $\mu$ and $\Sigma$ such that $q_\phi(z|x)= \mathcal{N}(\mu,\Sigma)$, from which a latent vector $z$ is generated via the reparametrization trick~\cite{kingmawelling}.

\subsection{Cross and Distribution Aligned VAE}

The goal of our model is to learn representations within a common space for a combination of $M$ data modalities. Hence, our model includes $M$ encoders, one for every modality, to map into this representation space. To minimize information loss, the original data must be reconstructed via the decoder networks. In effect, the basic VAE loss of our model is the sum of $M$ VAE-losses:
\begin{align}
\mathcal{L}_{VAE} = & \sum_i^M \mathbb{E}_{q_\phi(z|x)}[\log p_\theta(x^{(i)}|z)] \\
&- \beta D_{KL}( q_\phi(z|x^{(i)})||p_\theta(z) ) \nonumber
\end{align}
where $\beta$ weights the KL-Divergence~\cite{higgins2016beta}.
In the case of matching image features with class embeddings, $M=2$, $x^{(1)} \in X$ and $x^{(2)} \in C(Y^S)$. Yet, making the modality-specific autoencoders learn similar representations across modalities requires additional regularization terms. Therefore, our model aligns the latent distributions explicitly and enforces a cross-reconstruction criterion. In Figure~\ref{fig:crossvae} we show an overview of our model, depicting these two forms of latent distribution matching, which we refer to as Cross-Alignment (CA) and Distribution-Alignment (DA).

\myparagraph{Cross-Alignment (CA) Loss.}
Here, reconstructions are obtained by decoding the latent encoding of a sample from another modality, but the same class. Hence, every modality-specific decoder is trained on the latent vectors derived from the other modalities. This cross-reconstruction loss is:
\begin{equation}
\mathcal{L}_{CA} = \sum_i^M \sum_{j\neq i}^M 
|x^{(j)} - D_j(E_i(x^{(i)}))|.
\end{equation}
where $E_i$ is the encoder of a feature of $i^\mathrm{th}$ modality and $D_j$ is the decoder of a feature of the same class but the $j^\mathrm{th}$ modality.

\myparagraph{Distribution-Alignment (DA) Loss.}
Generated image and class representations can also be matched by minimizing their distance.
Here, we minimize the Wasserstein distance between the latent multivariate Gaussian distributions. In the case of multivariate Gaussians, a closed-form solution of the $\mathrm{2}$-Wasserstein distance~\cite{givens1984class} between two distributions $i$ and $j$ is given as:  
\begin{align}
W_{ij} & =\bigl[||\mu_i-\mu_j||^2_2 \\
& + Tr(\Sigma_i) + Tr(\Sigma_j) - 2(\Sigma_i^\frac{1}{2} \Sigma_i \Sigma_j^\frac{1}{2})^\frac{1}{2} \bigr]^\frac{1}{2}.   \nonumber
\end{align}
Since the encoder predicts diagonal covariance matrices, which are commutative, this distance simplifies to:
\begin{equation}
W_{ij}=\bigl(||\mu_i-\mu_j||^2_2 + ||\Sigma_i^\frac{1}{2}-\Sigma_j^\frac{1}{2}||^2_{\mathrm{Frobenius}}\bigr)^\frac{1}{2}
\end{equation}
and the Distribution Alignment (DA) loss for a group an M-tuple is written as:
\begin{equation}
\mathcal{L}_{DA} = \sum_i^M \sum_{j\neq i}^M W_{ij}.
\end{equation}

\myparagraph{Cross- and Distribution Alignment (\texttt{CADA-VAE}) Loss.}
The cross- and distribution aligned VAE combines the basic VAE-loss with $\mathcal{L}_{CA}$ (\texttt{CA-VAE}) and  $\mathcal{L}_{DA}$ (\texttt{DA-VAE}):
\begin{equation}
 \mathcal{L}_{CADA-VAE} = \mathcal{L}_{VAE} + \gamma \mathcal{L}_{CA} + \delta \mathcal{L}_{DA}
\end{equation}
where  $\gamma$ and $\delta$ are the weighting factors of the cross alignment and the distribution alignment loss, respectively. We show in section \ref{sec:cub-analysis} that our model can learn shared multimodal embeddings of more than two modalities, without examples of all modalities being available for all classes.

\myparagraph{Implementation Details.}
\label{sec:impl}
All encoders and decoders are multilayer perceptrons with one hidden layer. More hidden layers degraded performance as CNN features and attributes are already very high-level representations.
We use $1560$ hidden units for the image feature encoder and $1660$ for the decoder. The attribute encoder and decoder have $1450$ and $660$ hidden units, respectively.

The latent embedding size is $64$. For ImageNet, we chose a size of $128$, and use two hidden layers of identical size for the encoder, with the number of hidden units specified above and the image feature decoder layers are of size $1160$ and $1660$, while the attribute decoder uses $460$ and $660$ units. The model is trained for $100$ epochs by stochastic gradient descent using the Adam optimizer~\cite{adam} and a batch size of $128$ for ImageNet and $50$ for all other datasets. Each training batch consists of pairs of CNN features and \textit{matching} attributes, from different seen classes. A pair of data always belongs to the same class.
After individual VAEs learn to encode features of only their specific datatype for some epochs, we also start to compute cross- and distribution alignment losses.  
$\delta$  is increased from epoch $6$ to epoch $22$ by a rate of $0.54$ per epoch, while  $\gamma$ is increased from epoch $21$ to $75$ by $0.044$ per epoch.
For the KL-divergence we use an annealing scheme~\cite{bowman2016generating}, in which we increase the weight $\beta$ of the KL-divergence by a rate of $0.0026$ per epoch until epoch $90$. A KL-annealing scheme serves the purpose of first letting the VAE learn ``useful" representations before they are ``smoothed" out, since the KL-divergence would be otherwise a very strong regularizer~\cite{bowman2016generating}.

We empirically found that it is useful to use a variant of the reparametrization trick~\cite{kingmawelling}, in which all dimensions of the noise vector are sampled from a single unimodal Gaussian. Also, using the L1 distance as reconstruction error appeared to yield slightly better results than L2.
After training, the VAE encoders transform the training and test set of the final linear classifier into the latent space~\footnote{code at \href{https://github.com/edgarschnfld/CADA-VAE-PyTorch}{https://github.com/edgarschnfld/CADA-VAE-PyTorch}}.

\section{Experiments}
\label{sec:exp}
We evaluate our framework on four widely used benchmark datasets CUB-200-2011~\cite{cub} (CUB), SUN attribute (SUN)~\cite{sun}, Animals with Attributes $1$ and $2$ (AWA1~\cite{DAP}, AWA2~\cite{goodbadugly}) for the GZSL and GFSL settings.
All image features for VAE training originate from the $2048$-dimensional final pooling layer of a $\mathrm{ResNet}$-$\mathrm{101}$. To avoid violating the zero-shot assumption, i.e. test classes need to be disjoint from the classes that $\mathrm{ResNet}$-$\mathrm{101}$ was trained with, we use the proposed training splits in~\cite{goodbadugly}.

Attributes serve as class embeddings when available. For CUB, we also use sentence embeddings extracted from 10 sentences annotated per image averaged per class~\cite{reed} and for ImageNet we used Word2Vec~\cite{mikolov2013distributed} embeddings provided by~\cite{sync}.
All hyperparameters were chosen on a validation set provided by \cite{goodbadugly}. We report the harmonic mean (H) between seen (S) and unseen (U) average per-class accuracy.

{
\setlength{\tabcolsep}{6pt}
\renewcommand{\arraystretch}{1.2}
\begin{center}
\centering
\begin{table}[t]
\centering
\begin{tabular}{ l ccc}
 Model & \textbf{S}  & \textbf{U}  & \textbf{H} \\
 \hline
\texttt{DA-VAE} & $48.1$ & $43.8$ & $45.8$ \\
\texttt{CA-VAE} & $52.6$ & $48.1$ & $50.2$ \\
\texttt{CADA-VAE} & $\mathbf{53.5}$ & $\mathbf{51.6}$ & $\mathbf{52.4}$ \\
\end{tabular}
\caption{Ablation study. We compare GZSL accuracy on
CUB for different multi-modal alignment objective func-
tions, i.e. \texttt{DA-VAE} (distribution aligned VAE)
, \texttt{CA-VAE} (cross-aligned VAE) 
and \texttt{CADA-VAE} (cross and distribution aligned VAE).}
\label{table:ablation}
\end{table}
\end{center}
}

\vspace{-10mm}
\subsection{Analyzing \textbf{\texttt{CADA-VAE}} in Detail on CUB} 
\label{sec:cub-analysis}
In this section, we analyze several building blocks of our proposed framework such as the model, the choice of class embeddings as well as the size and the number of latent embeddings generated by our model in the GZSL setting.

\myparagraph{Analyzing Model Variants.}
In this ablation study, we present the results of different objective functions and the corresponding VAE variants, \texttt{CA-VAE} (cross-aligned VAE), \texttt{DA-VAE} (distribution-aligned VAE) and \texttt{CADA-VAE} (cross- and distribution-aligned VAE) on the CUB dataset in GZSL setting. 

As shown in Table~\ref{table:ablation}, the cross-alignment objective noticably improves performance compared to distribution alignment ($50.2\%$ vs. $45.8\%$). This is due to the fact that both seen and unseen class accuracies increase, i.e. seen class accuracy increases by $4.5\%$ and unseen class accuracy increases by $4.3\%$, when we use cross alignment loss rather than the distribution alignment loss.
Moreover, combining distribution alignment and the cross-alignment objectives, i.e. \texttt{CADA-VAE}, increases the accuracy to $52.4\%$ that comes from adding the distribution alignment to the CA-VAE. 
Our ablation study shows that aligning both the latent representations and the latent spaces is complementary since their combination leads to the highest result on both seen, unseen classes and their harmonic mean.

\begin{figure}[t]
\centering
\includegraphics[width=0.49\linewidth, trim=10 14 10 10, clip]{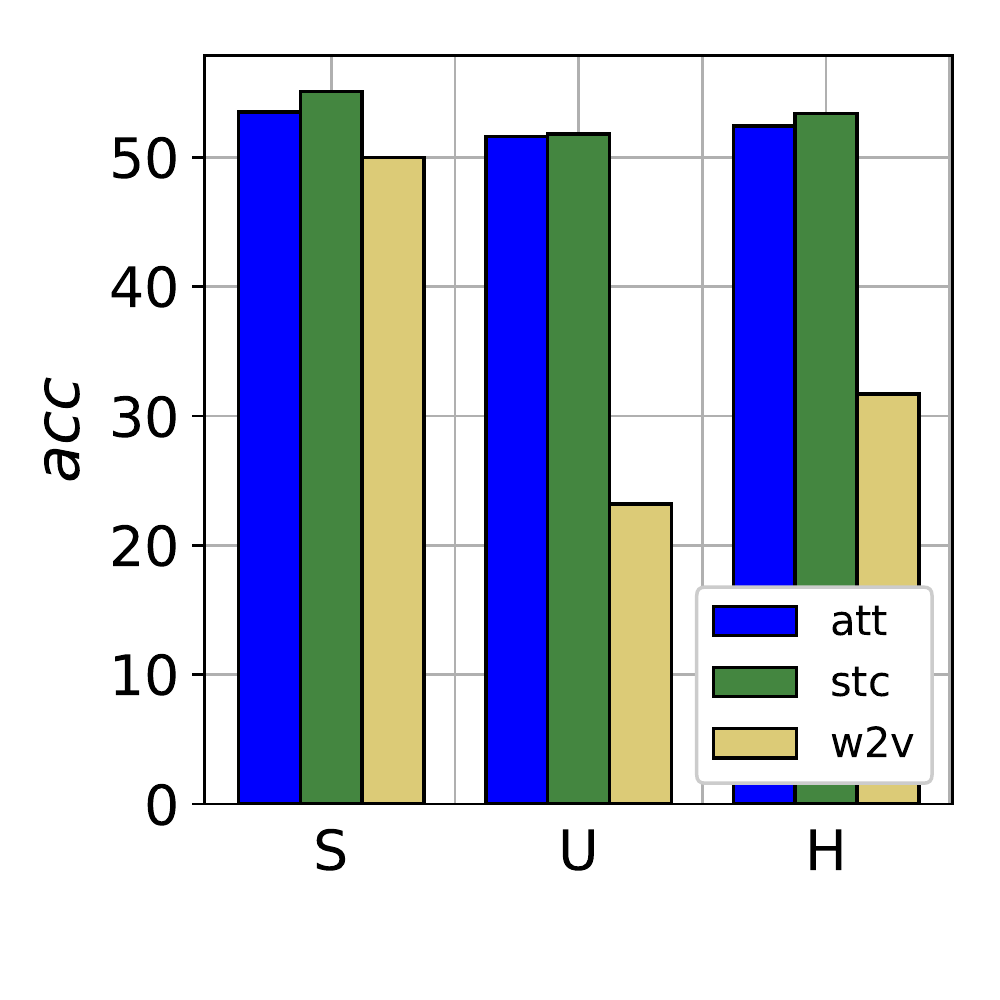}
\hfill
\includegraphics[width=0.49\linewidth, trim=10 14 10 10, clip]{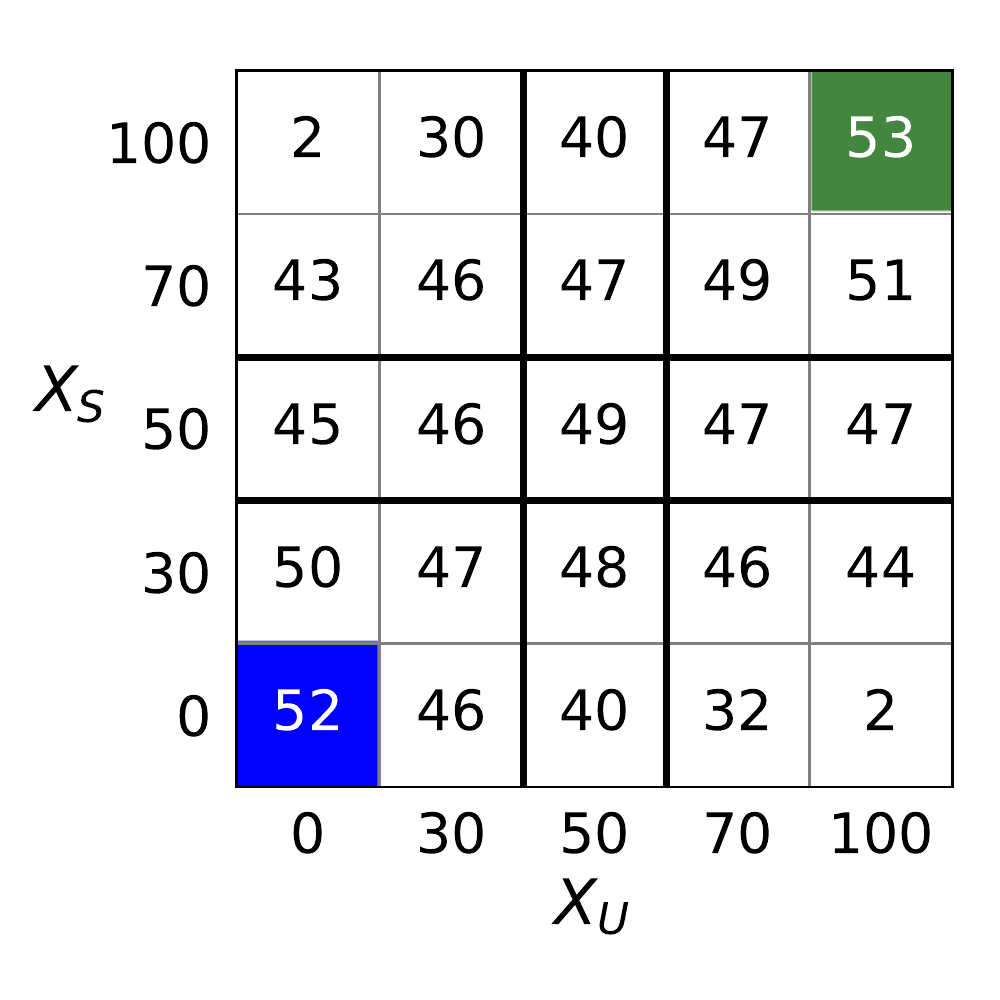}
\caption{Effect of different class embeddings. (Left) Seen, unseen and harmonic mean accuracy for CUB using different class embeddings as side information. (Right) Using both attributes and sentences as side information, i.e. $X_S$: the percentage of seen classes with sentences, $X_U$: the percentage of unseen classes with sentences. Attributes are the class embeddings for the $(100-X)\%$ of the classes.
}
\label{fig:other_embeddings}
\end{figure}

\myparagraph{Analyzing Side Information.} 
In sparse data regimes especially in zero-shot learning semantic representation of the classes, i.e. class embeddings, are as important as the image embeddings as they enable knowledge transfer from seen to unseen classes. We compare the results obtained with per-class attributes, per-class sentences and class-based Word2Vec representations.

Our results in Figure~\ref{fig:other_embeddings} (left) show that per-class sentence embeddings result in the best performance among all three, i.e. $53.4\%$, attributes follow closely, i.e. $52.4\%$.
With Word2Vec, the difference between the seen and the unseen class accuracy is large, indicating that the latent representations learned by Word2Vec are weak in robustness. This is expected, given that Word2Vec features do not explicitly or exclusively represent visual characteristics. In summary, these results demonstrate that our model is able to learn from various sources of side information. The results also show that latent features learned with more discriminative class embeddings lead to better overall accuracy.

To investigate one of the most prominent aspects of our model, i.e. the ability to handle missing side information, we train \texttt{CADA-VAE} such that $X_S \%$ of seen class image features are paired with sentence embeddings while the other $(100-X_S)\%$ of seen classes are paired with attributes. The setup is evaluated for $X_S=0,30,50,70,100$.
We also vary the fraction $X_U\%$ of unseen classes that are learned from sentence features (whereas $(100-X_U)\%$ denotes the fraction of unseen classes for which image features are only paired with attributes).

Figure \ref{fig:other_embeddings} (right) shows the results using different fractions of sentence embeddings and attribute embeddings for $X_S$ and $X_U$. When $X_U$ is held stable at 50\%, i.e. both seen and unseen classes have access to sentences and attributes equally half the time, we reach the highest accuracy for an $X_S$-$X_U$ ratio of $50\%$-$50\%$. 
Interestingly, at $(X_S=0,X_U=50)$, i.e. no seen-class sentences while unseen classes are represented by both attributes, the accuracy is $40\%$. On the other hand, at $(X_S=50,X_U=0)$, i.e. no unseen-class sentences while seen classes are represented by both attributes sentences, the accuracy increases to $45\%$. At $(X_S=50,X_U=100)$, i.e. 50\% attributes and 50\% sentences for seen classes but unseen classes are represented by sentences only, the accuracy further increases to $47\%$. 
These results indicate that sentences have an edge over attributes. However, when either sentences or attributes are not available, our model can recover the missing information from the other modality and still learn discriminative representations.

\begin{figure}[t]
\centering
\includegraphics[width=0.95\columnwidth]{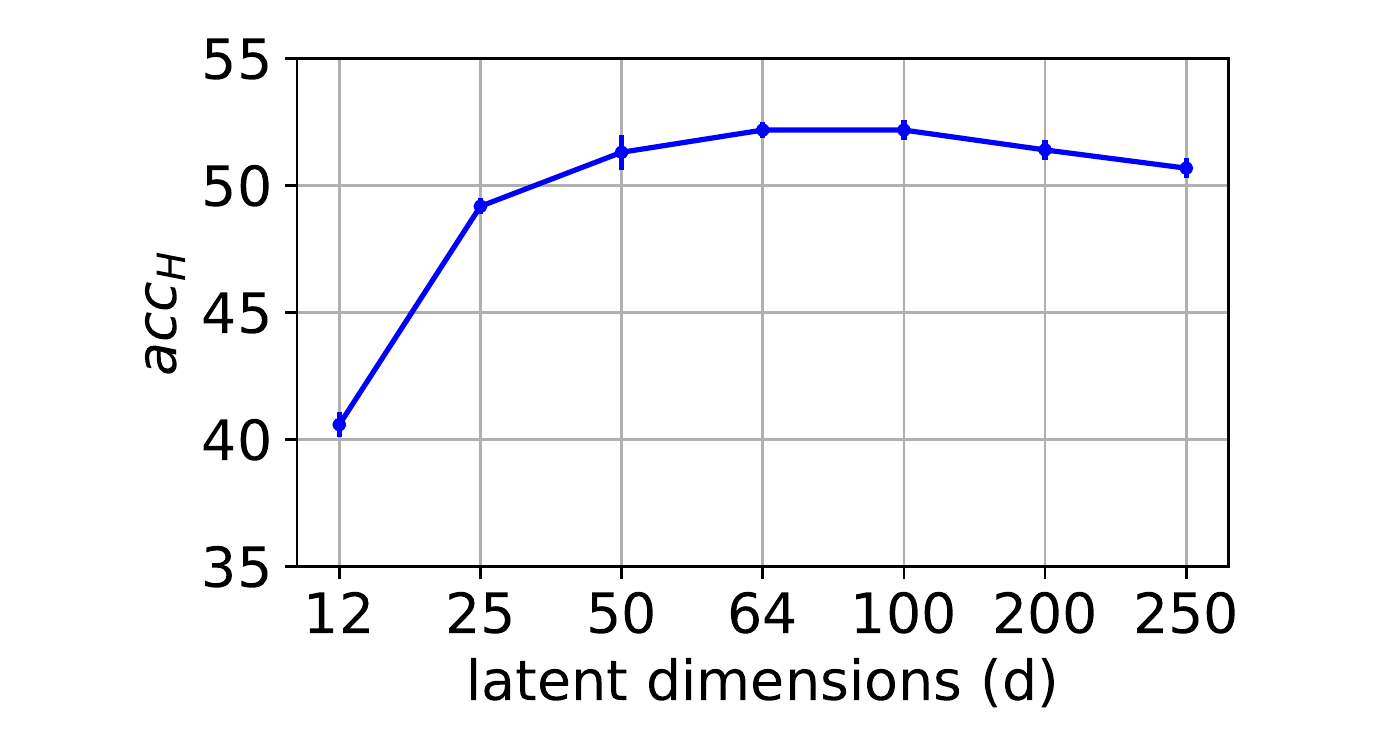}
\caption{The influence of the dimentionality of the latent features that are generated by \texttt{CADA-VAE} and used to train the GZSL classifier. We measure the harmonic mean accuracy on the CUB dataset}
\label{fig:latent_dimenstions}
\end{figure}

\myparagraph{Increasing Number of Latent Dimensions.}
In this analysis, we have explored the robustness of our method to the dimensionality of the latent space. Higher dimensions allow more degrees of freedom but require more data, while compact features capture the essential discriminative information. Without loss of generality, we report the harmonic mean accuracy of \texttt{CADA-VAE} for different latent dimensions on CUB, i.e. 
$12,25,50,64,100,200$ and $250$.

We observe in Figure \ref{fig:latent_dimenstions} that the accuracy initially increases with increasing dimensionality until it achieves its peak accuracy of $52.4\%$ at $d=64$ and flattens until $d=100$ after which it declines upon further increase of the latent dimension. We conclude from these experiments that the most discriminative properties of two modalities are captured when the latent space has around $64-100$ dimensions. For efficiency reasons, we use $64$ dimensional latent features for the rest of the paper.

\begin{figure}[t]
\centering
\includegraphics[width=0.48\linewidth, trim=0 10 20 10, clip]{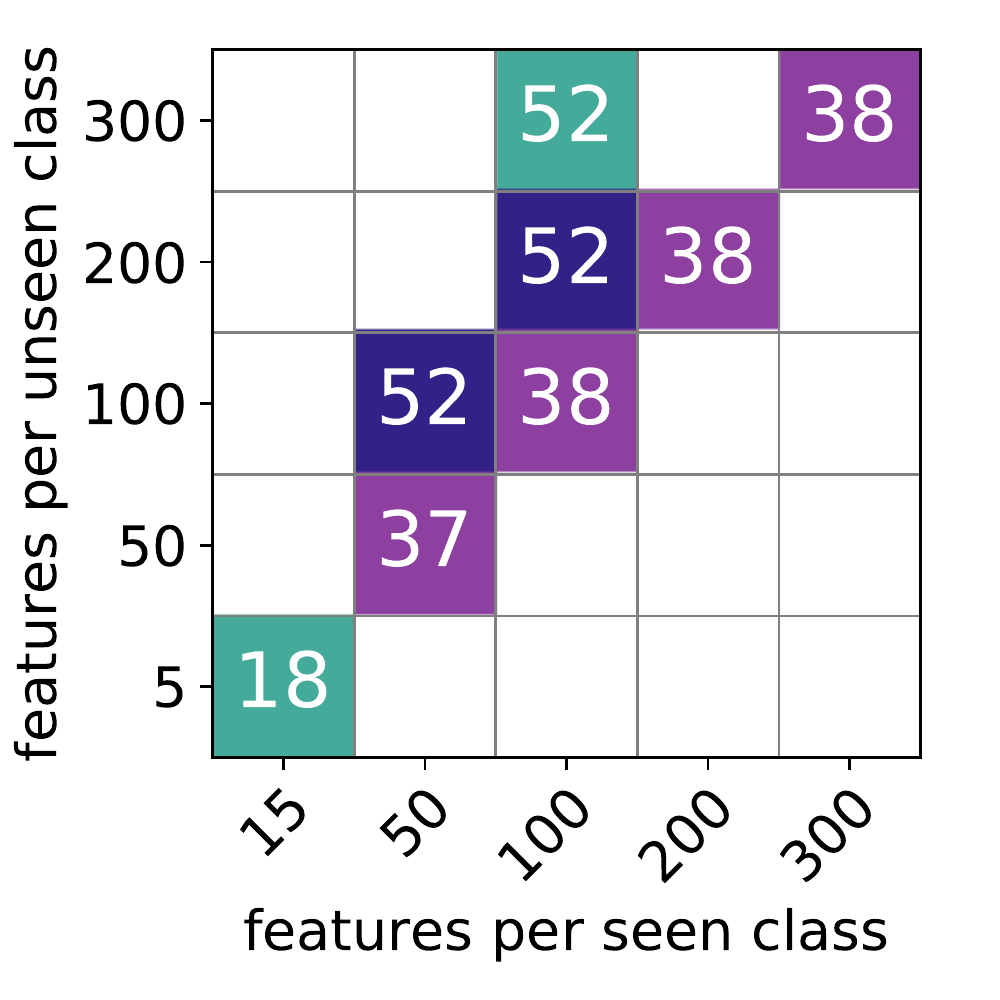}
\hfill
\includegraphics[width=0.48\linewidth,trim=0 10 20 10, clip]{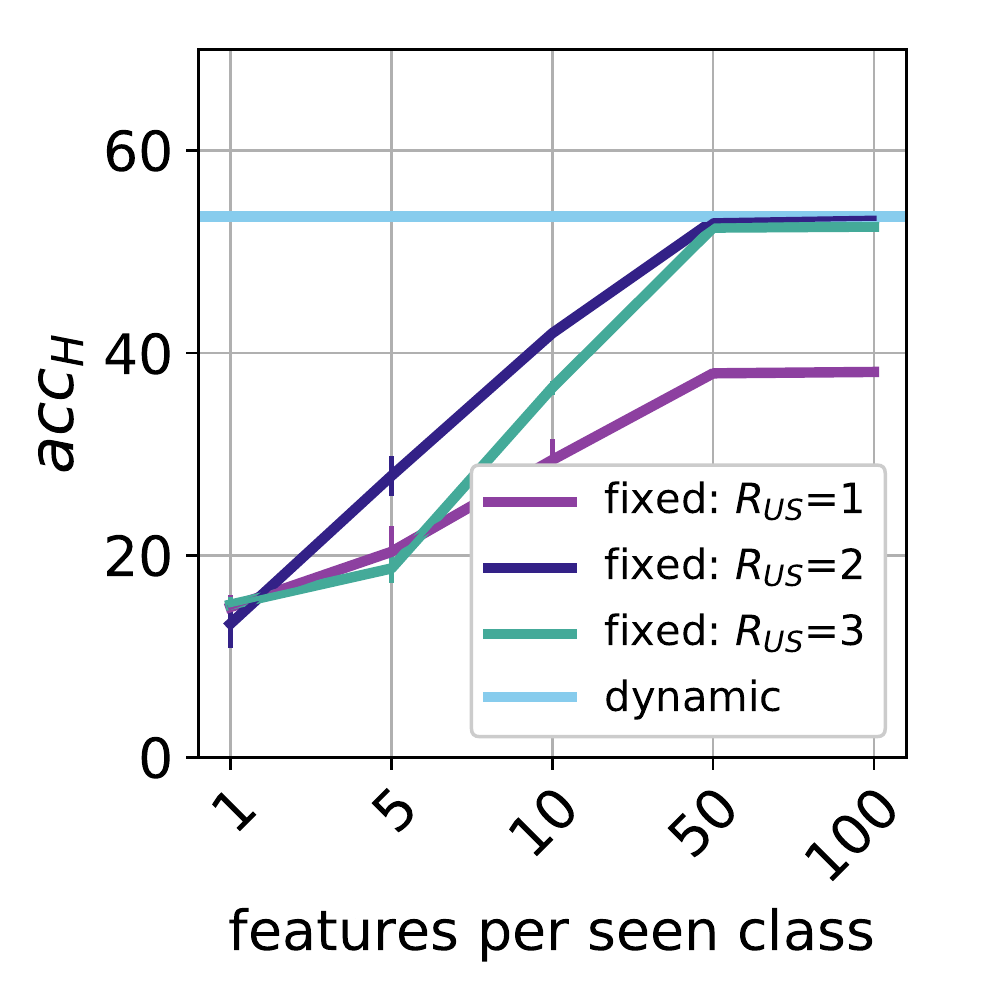}
\caption{Analyzing the effect of the number of latent features per class on the harmonic mean accuracy in GZSL. An unseen-seen ratio $R_{US}$ of 2 means that twice as many samples are generated for unseen classes than for seen classes . The dynamic dataset (light blue) does not rely on a fixed number of sampled latent features.
}
\label{fig:dynamic_dataset}
\end{figure}

\myparagraph{Increasing Number of Latent Features.}
Our model can be used to generate an arbitrarily large number of latent features. In this experiment, we vary the number of latent features per class from $1$ to $300$ on CUB in the GZSL setting and reach the optimum performance with $50$ or more latent features per seen class (Figure \ref{fig:dynamic_dataset}, left). In principle, seen and unseen classes do not need to have the same number of samples.
We also vary the number of features per seen and unseen classes. Indeed, the best accuracy is achieved when there are approximately twice as many features per unseen than seen classes which improves the accuracy from $37\%$ to $52\%$. While $100$ latent features per every class, i.e. $200\times 100 =20K$, gives $38\%$ accuracy, having $50$ latent features per seen classes and $100$ latent features per unseen class, i.e. $100\times50 + 50\times150 = 12.5K$, leads to $52\%$ accuracy. Hence, generating more features of under-represented classes is important for better accuracy.

As for our results in Figure \ref{fig:dynamic_dataset} on the right, we build a dynamic training set by generating latent features continuously at every iteration and do not use any sample more than once. Hence, we eliminate one of the tunable parameters, i.e. the number of latent features to generate. Because of the non-deterministic mapping of the VAE encoder, every latent feature of a different class is unique. Our results indicate that the best accuracy is achieved when unseen and seen class samples are equally balanced. In CUB, using a dynamic training set reaches the same performance as using a fixed dataset with $100$ unseen examples and $50$ seen examples. On the other hand, using a fixed dataset leads to a faster training procedure. Hence, we use a fixed dataset with $200$ examples per seen class and $400$ examples per unseen class in every benchmark reported in this paper.

{
\setlength{\tabcolsep}{5pt}
\renewcommand{\arraystretch}{1.2}
\begin{table*}[t]
\begin{center}
\resizebox{\textwidth}{!}{%
\begin{tabular}{ l c  ccc ccc ccc ccc }
  & &  \multicolumn{3}{c}{CUB }  & \multicolumn{3}{c}{SUN }  & \multicolumn{3}{c}{AWA1} & \multicolumn{3}{c}{AWA2}  \\
 Model & Feature Size & \textbf{S}   & \textbf{U}   & \textbf{H}   & \textbf{S}   & \textbf{U}   & \textbf{H}
 & \textbf{S}   & \textbf{U}   & \textbf{H}  & \textbf{S}   & \textbf{U}   & \textbf{H} \\
 \hline
CMT \cite{socher2013zero} & \multirow{7}{*}{$2048$} & $49.8$ & $7.2$ & $12.6$ & $21.8$ & $8.1$ & $11.8$ & $87.6$ & $0.9$ & $1.8$ & $90.0$ & $0.5$ & $1.0$
\\
SJE \cite{sje}  & & $59.2$ & $23.5$ & $33.6$ & $30.5$ & $14.7$ & $19.8$ & $74.6$ & $11.3$ & $19.6$ & $73.9$ & $8.0$ & $14.4$
\\
ALE \cite{ss} & & $62.8$ & $23.7$ & $34.4$ & $33.1$ & $21.8$ & $26.3$ & $76.1$ & $16.8$ & $27.5$ & $81.8$ & $14.0$ & $23.9$
\\
LATEM \cite{latem} & &  $57.3$ & $15.2$ & $24.0$ & $28.8$ & $14.7$ & $19.5$ & $71.7$ & $7.3$ & $13.3$ & $77.3$ & $11.5$ & $20.0$
\\
EZSL \cite{ezsl} & & $63.8$ & $12.6$ & $21.0$ & $27.9$ & $11.0$ & $15.8$ & $75.6$ & $6.6$ & $12.1$ & $77.8$ & $5.9$ & $11.0$
\\
SYNC \cite{sync} & & $70.9$ & $11.5$ & $19.8$ & $43.3$ & $7.9$ & $13.4$ & $87.3$ & $8.9$ & $16.2$ & $90.5$ & $10.0$ & $18.0 $
\\
DeViSE \cite{devise}  &  & $53.0$ & $23.8$ & $32.8$ & $27.4$ & $16.9$ & $20.9$ & $68.7$ & $13.4$ & $22.4$ & $74.7$ & $17.1$ & $27.8 $
\\
\hline
f-CLSWGAN \cite{featgen}& \multirow{3}{*}{$1024$} & $57.7$ & $43.7$ & $49.7$ & $36.6$ & $42.6$ & $39.4$ & $61.4$ & $57.9$ & $59.6 $ & $ 68.9$ & $52.1 $ & $59.4$
\\
CVAE \cite{cvae} & & -- & -- & $34.5$ & -- & -- & $26.7$ & -- & -- & $47.2$ & -- & -- & $51.2$
\\
SE \cite{segzsl}  & & $53.3$ & $41.5$ & $46.7$ & $30.5$ & $40.9$ & $34.9$ & $67.8$ & $56.3$ & $61.5$ & $68.1$ & $58.3$ & $62.8$
\\
\hline
ReViSE \cite{tsai2017learning} & $75/100$ & $28.3$ & $37.6$ & $32.3$ & $20.1$ & $24.3$ & $22.0$ & $37.1$ & $46.1$ & $41.1$ & $39.7$ & $46.4$ & $42.8$
\\
\hline
ours (\texttt{CADA-VAE}) & $64$ & $53.5$ & $51.6$ & $\mathbf{52.4}$  & $35.7$ & $47.2$ & $\mathbf{40.6}$ & $72.8$ & $57.3$ & $\mathbf{64.1}$ & $75.0$ & $55.8$ & $\mathbf{63.9}$
\end{tabular}
}
\end{center}
\vspace{-3mm}
\caption{Comparing \texttt{CADA-VAE} with the state of the art. We report per class accuracy for seen (S) and unseen (S) classes and their harmonic mean (H). All reported numbers for our method are averaged over ten runs.}
\label{table:gzsl_scores}
\end{table*}
}

\subsection{Comparing \textbf{\texttt{CADA-VAE}} on Benchmark Datasets}
In this section, we compare our \texttt{CADA-VAE} on four benchmark datasets, i.e. CUB, SUN, AWA1 and AWA2, in the GZSL and GFSL setting.

\myparagraph{Generalized Zero-Shot Learning.}
We compare our model with 11 state-of-the-art models. Among those, CVAE~\cite{cvae}, SE~\cite{segzsl}, and f-CLSWGAN~\cite{featgen} learn to generate artificial visual data and thereby treat the zero-shot learning task as a data-augmentation task.
On the other hand, the classic ZSL methods DeViSE~\cite{devise}, SJE~\cite{sje}, ALE~\cite{ss}, EZSL~\cite{ezsl} and LATEM~\cite{latem} use a linear compatibility function or other similarity metrics to compare embedded visual and semantic features; CMT~\cite{socher2013zero} and LATEM~\cite{latem} utilize multiple neural networks to learn a non-linear embedding; and SYNC~\cite{sync} aligns a class embedding space and a weighted bipartite graph. ReViSE~\cite{tsai2017learning} learns a shared latent manifold between the image features and class attributes using autoencoders. 

The results in Table \ref{table:gzsl_scores} show that our \texttt{CADA-VAE} outperforms all other methods on all datasets. The accuracy difference between our model and the closest baseline, ReViSE~\cite{tsai2017learning}, is as follows: $52.4\%$ vs $32.3\%$ on CUB, $40.6\%$ vs $22.0\%$ on SUN, $64.1\%$ vs $41.1\%$ on AWA1 and $63.9\%$ vs $42.8\%$ on AWA2. Moreover, our model achieves significant improvements over feature generating models most notably on CUB. In doing so, \texttt{CADA-VAE} is the first cross-modal embedding model to outperform methods based on feature-augmentation. Compared to the classic methods, our model leads to at least $100\%$ improvement in harmonic mean accuracies. In the legacy challenge of zero-shot learning, \texttt{CADA-VAE} provides competitive performance, i.e. $60.4$ on CUB, $61.8$ on SUN, $62.3$ on AWA1, $64.0$ on AWA2. However, in this work, we focus on the more practical and challenging GZSL setting.

Since our model does not use any CNN features, i.e. we generate $64$-dimensional latent features for all classes, it achieves a balance between seen and unseen class accuracies better than CNN feature-generating approaches especially on CUB.
In addition, \texttt{CADA-VAE} learns
a shared representation in a weakly-supervised fashion, through a cross-reconstruction objective. Since the latent features have to be decoded into every involved modality, and since every modality encodes complementary information, the model is encouraged to learn an encoding that retains the information contained in all used modalities. In doing so, our method is less biased towards learning the distribution of the seen class image features, which is known as the projection domain shift problem~\cite{fu2014transductive}. As we generate a certain number of latent features per class using non-deterministic encoders, our method is also akin to data-generating approaches. However, the learned representations lie in a lower dimensional space, i.e. only $64$, and therefore, are less prone to bias towards the training set of image features. 
In effect, our training is more stable than the adversarial training schemes used for data generation~\cite{featgen}. In fact, we did not conduct any dataset specific parameter tuning and use the same parameters for all datasets.

\begin{figure*}[t]
    \includegraphics[width=0.24\textwidth, trim=5 0 10 0, clip]{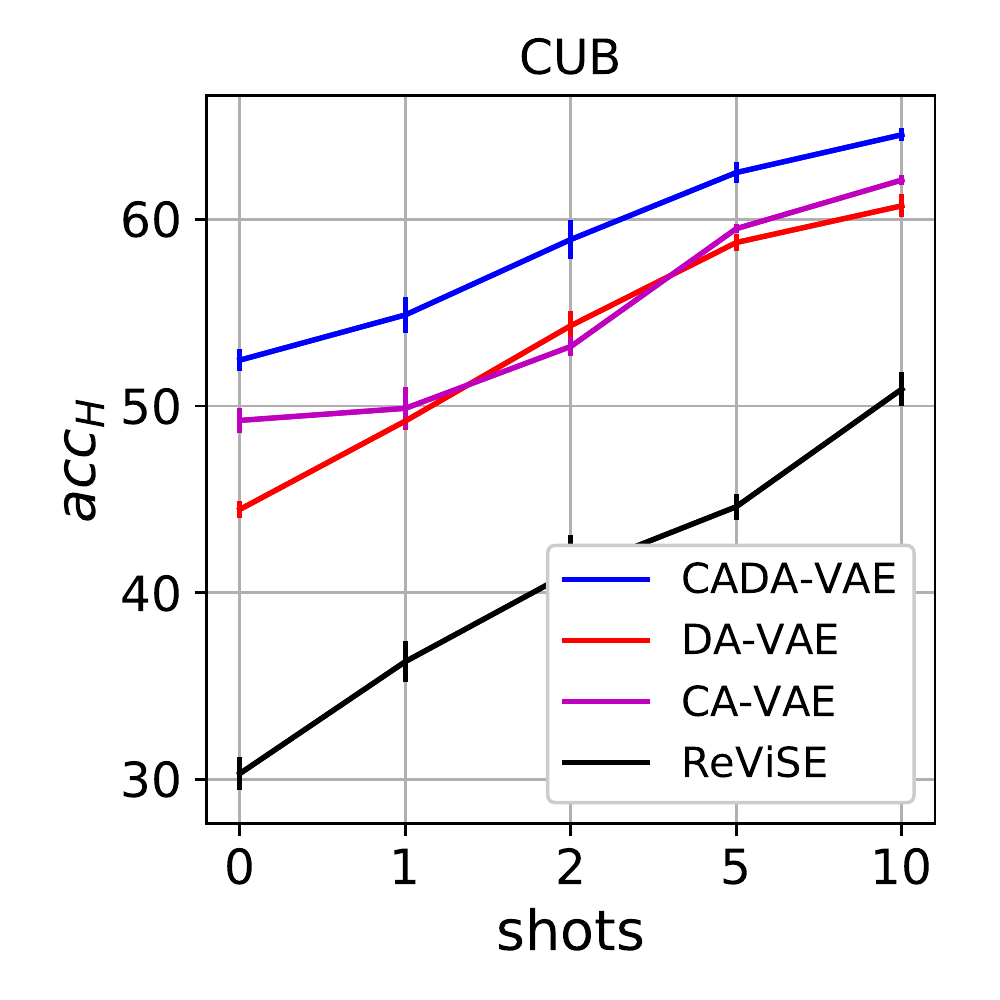}
    \includegraphics[width=0.24\textwidth, trim=5 0 10 0, clip]{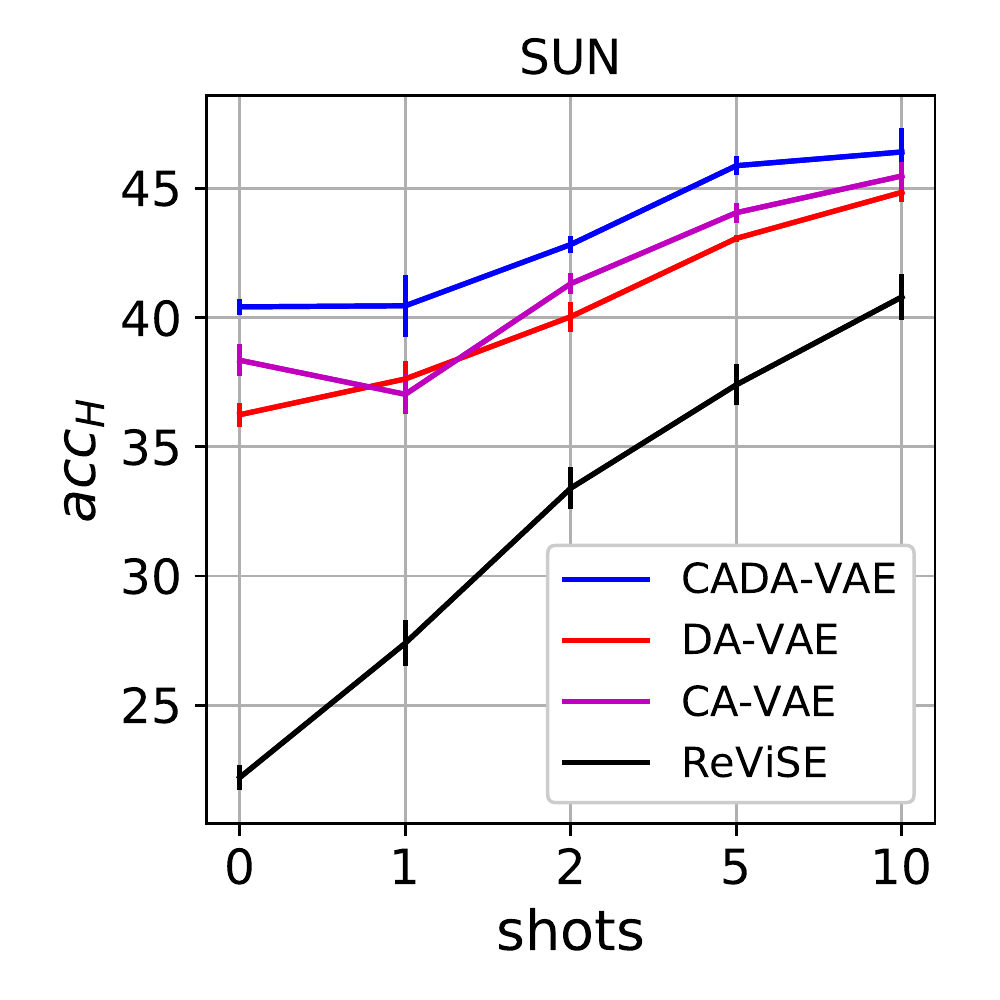}
    \includegraphics[width=0.24\textwidth, trim=5 0 10 0, clip]{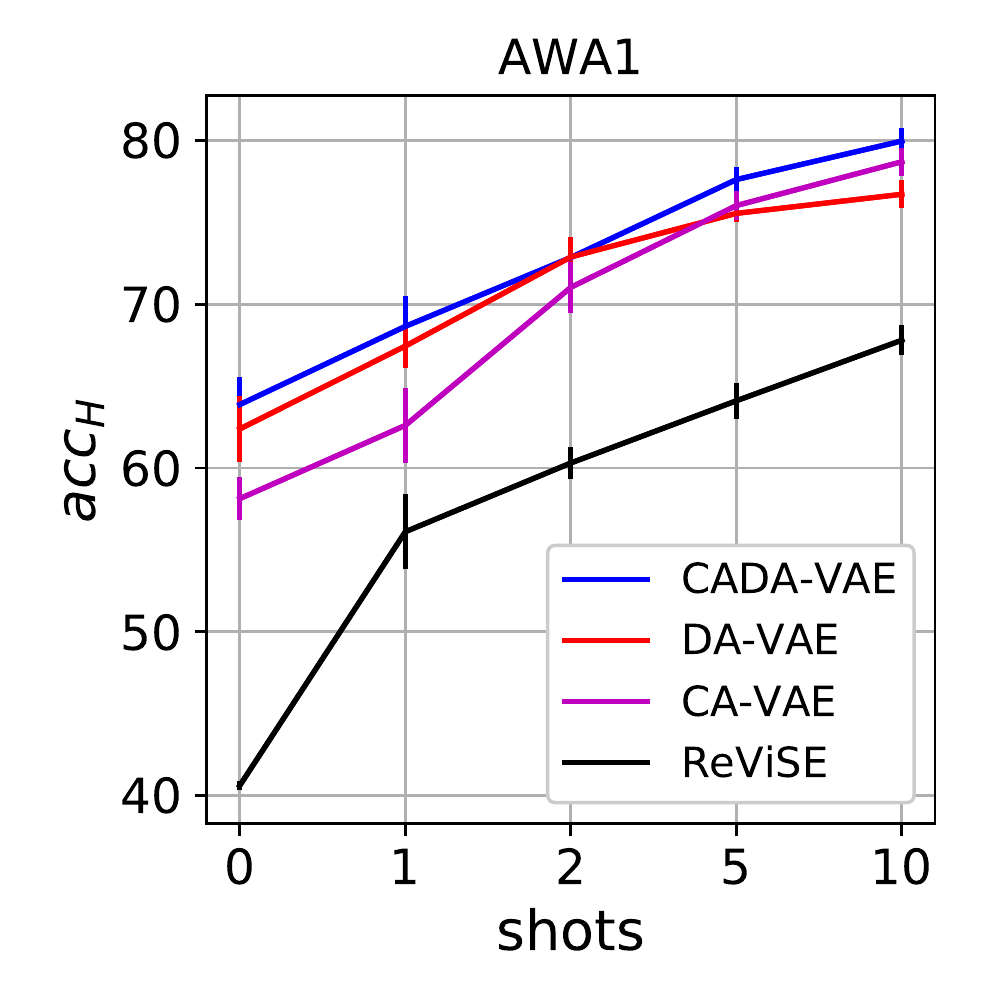}
    \includegraphics[width=0.24\textwidth, trim=5 0 10 0, clip]{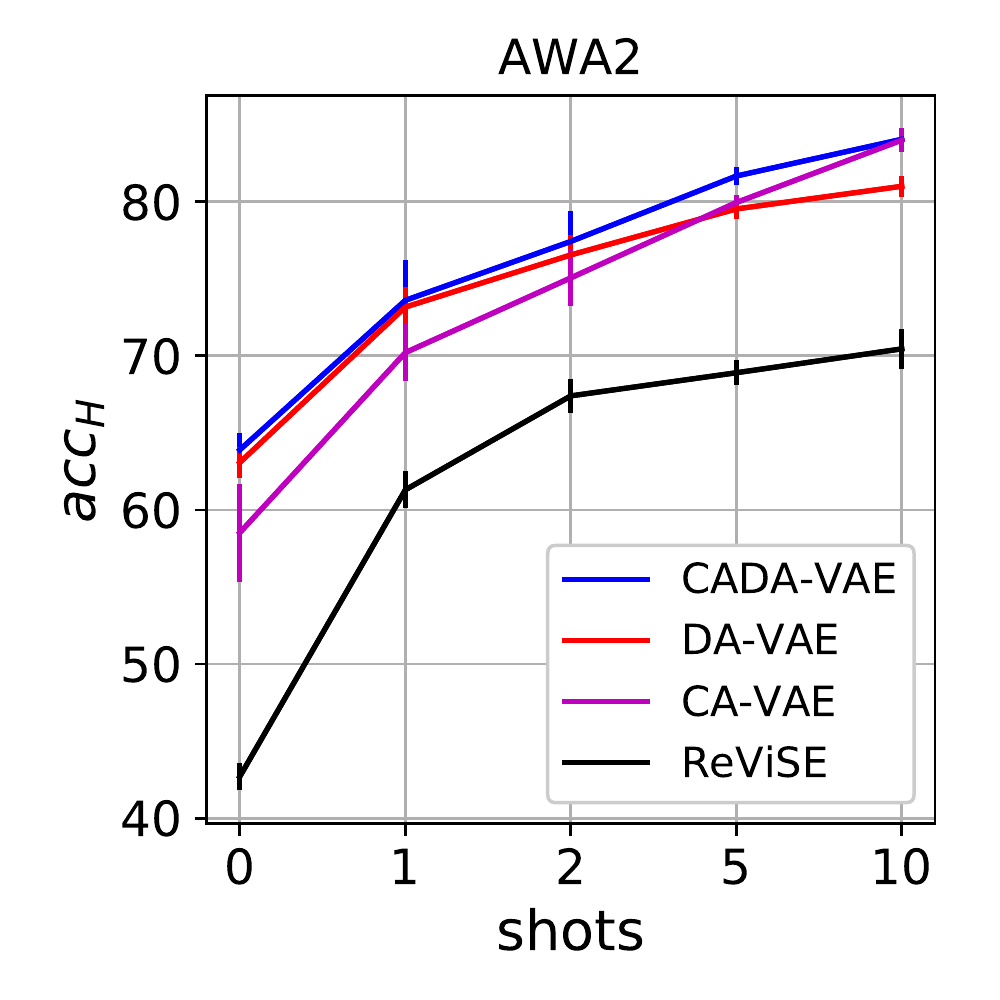}
    \vspace{-3mm}
\caption{Comparing \texttt{CA-VAE}, \texttt{DA-VAE}, \texttt{CADA-VAE} with ReViSE \cite{tsai2017improving} with increasing numbers of training samples from unseen classes, i.e. in the generalized few-shot setting.}
\label{fig:GFSL}
\end{figure*}

\myparagraph{Generalized Few-Shot Learning.}
We evaluate our models by using zero, two, five and ten shots for GFSL on four datasets. We compare our results with the most similar published work in this domain, i.e. ReViSE~\cite{tsai2017improving}. 
Figure \ref{fig:GFSL} shows that our latent representations learned from the side information improves over the GZSL setting significantly even by including only a few labeled samples. Specifically, adding a single latent feature from unseen classes to the training set improves the accuracy by $1$-$10\%$, depending on the dataset. While on CUB the accuracy improvement from $0$ to $10$ shots is $12\%$, on AWA1\&2 this improvement reaches $20\%$. Moreover, while the harmonic mean accuracy increases with the number of shots in both methods, all variants of our method outperform the baseline by a large margin across all the datasets indicating the generalization capability of our method to the GFSL setting.

Furthermore, similar to the GZSL scenario, on the fine-grained CUB and SUN datasets, \texttt{CADA-VAE} reaches the highest performance where it is followed by \texttt{CA-VAE} and \texttt{DA-VAE}, respectively. However,
on AWA1 and AWA2 the difference between different models is not significant. We associate this with the fact that as AWA1 and AWA2 datasets are coarse-grained datasets, the image features are already discriminative. Hence, aligning the latent space with attributes does not lead to a significant difference.

\begin{figure}[t]
\centering
\includegraphics[width=1.0\linewidth, trim=10 0 70 30,clip]{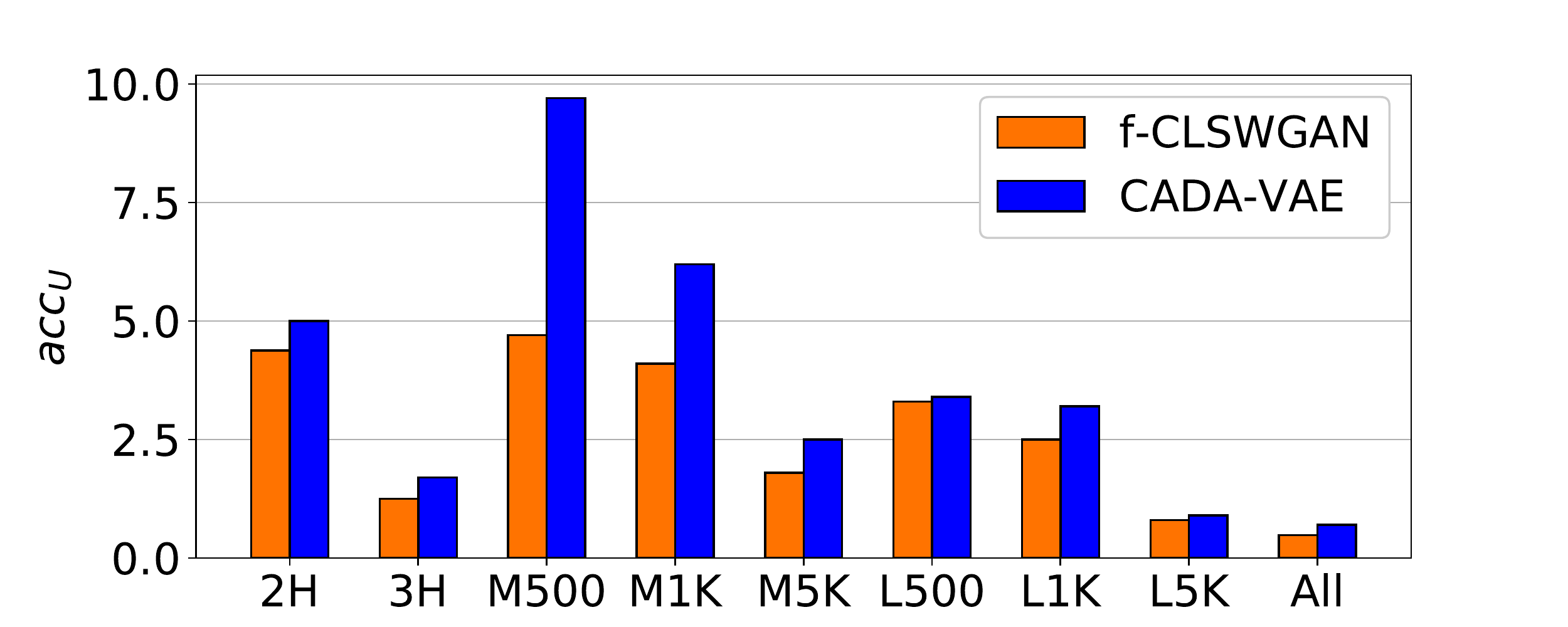}
\caption{ImageNet results on GZSL. We report the top-1 accuracy for unseen classes. Both \texttt{f-CLSWGAN} and \texttt{CADA-VAE} use a linear softmax classifier.}
\label{fig:imagenet_results}
\end{figure}

\subsection{ImageNet Experiments}

The ImageNet dataset serves as a challenging testbed for GZSL. In~\cite{goodbadugly} several evaluation splits were proposed with increasing granularity and size both in terms of the number of classes and the number of images. Note that since all the images of $1\mathrm{K}$ classes are used to train $\mathrm{ResNet}$-$\mathrm{101}$, measuring seen class accuracies would be biased. However, we can still evaluate the accuracy of unseen class images in the GZSL search space that contains both seen and unseen classes. Hence, at test time the $1K$ seen classes act as distractors. This way, we can measure the transferability of our latent representations to completely unseen classes, i.e. classes that are not seen either during ResNet training nor \texttt{CADA-VAE} training. For ImageNet, since attributes are not available, we use Word2Vec features as class embeddings provided by~\cite{sync}. 
We compare our model with \texttt{f-CLSWGAN}~\cite{featgen}, i.e. an image feature generating framework which currently achieves the state of the art on ImageNet. We use the same evaluation protocol on all the splits. Among the splits, 2H and 3H are the classes $2$ or $3$ hops away from the $1K$ seen training classes of ImageNet according to the ImageNet hierarchy. M500, M1K and M5K are the $500$, $1000$ and $5000$ most populated classes, while L500, L1K and L5K are the $500$, $1000$ and $5000$ least populated classes that come from the rest of the $21\mathrm{K}$ classes. Finally, `All` denotes the remaining $20K$ classes.

As shown in Figure~\ref{fig:imagenet_results}, our model significantly improves the state of the art in all splits. The accuracy improvement is significant especially on M500 and M1K splits, i.e. for M500 the search space is $1.5K$ classes, for M1K, the search space consists of $2K$ classes. For the $L500$, $L1K$ and $L5K$ splits, there are on average only $1$, $3$ and $5$ images per class available~\cite{goodbadugly}. Since the test time search space in the `All` split is $22\mathrm{K}$ dimensional, even a small improvement in accuracy is considered to be compelling. The achieved substantial increase in performance by \texttt{CADA-VAE} shows that our $128$-dim latent feature space constitutes a robust generalizable representation, surpassing the current state-of-the-art image feature generating framework \texttt{f-CLSWGAN}.

\section{Conclusion}
\label{sec:con}

In this work, we propose \texttt{CADA-VAE}, a cross-modal embedding framework for generalized zero- and few-shot learning. In \texttt{CADA-VAE}, we train a VAE for both visual and semantic modalities. The VAE of each modality has to jointly represent the information embodied by all modalities in its latent space. The corresponding latent distributions are aligned by minimizing their Wasserstein distance and by enforcing cross-reconstruction. This procedure leaves us with encoders that can encode features from different modalities into one cross-modal embedding space, in which a linear softmax classifier can be trained.
We present different variants of cross-aligned and distribution aligned VAEs and establish new state-of-the-art results in generalized zero-shot learning for four medium-scale benchmark datasets as well as the large-scale ImageNet. We further show that a cross-modal embedding model for generalized zero-shot learning achieves better performance than data-generating methods, establishing the new state of the art.

{\small
\bibliographystyle{ieee}
\bibliography{egbib}

\begin{thebibliography}{10}\itemsep=-1pt

\bibitem{ss}
Z.~Akata, F.~Perronnin, Z.~Harchaoui, and C.~Schmid.
\newblock Label-embedding for image classification.
\newblock {\em TPAMI}, 38(7):1425--1438, 2016.

\bibitem{sje}
Z.~Akata, S.~Reed, D.~Walter, H.~Lee, and B.~Schiele.
\newblock Evaluation of output embeddings for fine-grained image
  classification.
\newblock In {\em CVPR}, pages 2927--2936, 2015.

\bibitem{bowman2016generating}
S.~R. Bowman, L.~Vilnis, O.~Vinyals, A.~Dai, R.~Jozefowicz, and S.~Bengio.
\newblock Generating sentences from a continuous space.
\newblock In {\em CoNLL}, pages 10--21, 2016.

\bibitem{sync}
S.~Changpinyo, W.-L. Chao, B.~Gong, and F.~Sha.
\newblock Synthesized classifiers for zero-shot learning.
\newblock In {\em CVPR}, pages 5327--5336, 2016.

\bibitem{deng2009imagenet}
J.~Deng, W.~Dong, R.~Socher, L.-J. Li, K.~Li, and L.~Fei-Fei.
\newblock Imagenet: A large-scale hierarchical image database.
\newblock In {\em CVPR}, 2009.

\bibitem{devise}
A.~Frome, G.~S. Corrado, J.~Shlens, S.~Bengio, J.~Dean, T.~Mikolov, et~al.
\newblock Devise: A deep visual-semantic embedding model.
\newblock In {\em NIPS}, pages 2121--2129, 2013.

\bibitem{fu2014transductive}
Y.~Fu, T.~M. Hospedales, T.~Xiang, Z.~Fu, and S.~Gong.
\newblock Transductive multi-view embedding for zero-shot recognition and
  annotation.
\newblock In {\em ECCV}, pages 584--599, 2014.

\bibitem{givens1984class}
C.~R. Givens, R.~M. Shortt, et~al.
\newblock A class of wasserstein metrics for probability distributions.
\newblock {\em The Michigan Mathematical Journal}, 31(2):231--240, 1984.

\bibitem{gretton2012kernel}
A.~Gretton, K.~M. Borgwardt, M.~J. Rasch, B.~Sch{\"o}lkopf, and A.~Smola.
\newblock A kernel two-sample test.
\newblock {\em JMLR}, 13(Mar):723--773, 2012.

\bibitem{hallucinating}
B.~Hariharan and R.~B. Girshick.
\newblock Low-shot visual recognition by shrinking and hallucinating features.
\newblock In {\em ICCV}, pages 3037--3046, 2017.

\bibitem{higgins2016beta}
I.~Higgins, L.~Matthey, A.~Pal, C.~Burgess, X.~Glorot, M.~Botvinick,
  S.~Mohamed, and A.~Lerchner.
\newblock beta-vae: Learning basic visual concepts with a constrained
  variational framework.
\newblock 2016.

\bibitem{adam}
D.~P. Kingma and J.~Ba.
\newblock Adam: A method for stochastic optimization.
\newblock In {\em ICLR}, 2015.

\bibitem{kingmawelling}
D.~P. Kingma and M.~Welling.
\newblock Auto-encoding variational bayes.
\newblock In {\em ICLR}, 2014.

\bibitem{segzsl}
V.~Kumar~Verma, G.~Arora, A.~Mishra, and P.~Rai.
\newblock Generalized zero-shot learning via synthesized examples.
\newblock In {\em CVPR}, pages 4281--4289, 2018.

\bibitem{DAP}
C.~H. Lampert, H.~Nickisch, and S.~Harmeling.
\newblock Learning to detect unseen object classes by between-class attribute
  transfer.
\newblock In {\em CVPR}, pages 951--958, 2009.

\bibitem{imagetranslation}
M.-Y. Liu, T.~Breuel, and J.~Kautz.
\newblock Unsupervised image-to-image translation networks.
\newblock In {\em NIPS}, pages 700--708, 2017.

\bibitem{mikolov2013distributed}
T.~Mikolov, I.~Sutskever, K.~Chen, G.~S. Corrado, and J.~Dean.
\newblock Distributed representations of words and phrases and their
  compositionality.
\newblock In {\em NIPS}, pages 3111--3119, 2013.

\bibitem{cvae}
A.~Mishra, S.~Krishna~Reddy, A.~Mittal, and H.~A. Murthy.
\newblock A generative model for zero shot learning using conditional
  variational autoencoders.
\newblock In {\em CVPR}, pages 2188--2196, 2018.

\bibitem{DMAE}
T.~Mukherjee, M.~Yamada, and T.~M. Hospedales.
\newblock Deep matching autoencoders.
\newblock {\em arXiv preprint arXiv:1711.06047}, 2017.

\bibitem{norouzi2013zero}
M.~Norouzi, T.~Mikolov, S.~Bengio, Y.~Singer, J.~Shlens, A.~Frome, G.~S.
  Corrado, and J.~Dean.
\newblock Zero-shot learning by convex combination of semantic embeddings.
\newblock In {\em ICLR}, 2014.

\bibitem{sun}
G.~Patterson and J.~Hays.
\newblock Sun attribute database: Discovering, annotating, and recognizing
  scene attributes.
\newblock In {\em CVPR}, pages 2751--2758, 2012.

\bibitem{ramakrishnan2017empirical}
S.~K. Ramakrishnan, A.~Pal, G.~Sharma, and A.~Mittal.
\newblock An empirical evaluation of visual question answering for novel
  objects.
\newblock In {\em CVPR}, pages 4392--4401, 2017.

\bibitem{reed}
S.~Reed, Z.~Akata, H.~Lee, and B.~Schiele.
\newblock Learning deep representations of fine-grained visual descriptions.
\newblock In {\em CVPR}, pages 49--58, 2016.

\bibitem{ezsl}
B.~Romera-Paredes and P.~Torr.
\newblock An embarrassingly simple approach to zero-shot learning.
\newblock In {\em ICML}, pages 2152--2161, 2015.

\bibitem{styletransfer}
T.~Shen, T.~Lei, R.~Barzilay, and T.~Jaakkola.
\newblock Style transfer from non-parallel text by cross-alignment.
\newblock In {\em NIPS}, pages 6830--6841, 2017.

\bibitem{snell2017prototypical}
J.~Snell, K.~Swersky, and R.~Zemel.
\newblock Prototypical networks for few-shot learning.
\newblock In {\em NIPS}, pages 4077--4087, 2017.

\bibitem{socher2013zero}
R.~Socher, M.~Ganjoo, C.~D. Manning, and A.~Ng.
\newblock Zero-shot learning through cross-modal transfer.
\newblock In {\em NIPS}, pages 935--943, 2013.

\bibitem{smi}
T.~Suzuki and M.~Sugiyama.
\newblock Sufficient dimension reduction via squared-loss mutual information
  estimation.
\newblock In {\em AIStats}, pages 804--811, 2010.

\bibitem{tsai2017learning}
Y.-H.~H. Tsai, L.-K. Huang, and R.~Salakhutdinov.
\newblock Learning robust visual-semantic embeddings.
\newblock In {\em ICCV}, pages 3591--3600, 2017.

\bibitem{tsai2017improving}
Y.-H.~H. Tsai and R.~Salakhutdinov.
\newblock Improving one-shot learning through fusing side information.
\newblock {\em arXiv preprint arXiv:1710.08347}, 2017.

\bibitem{adda}
E.~Tzeng, J.~Hoffman, K.~Saenko, and T.~Darrell.
\newblock Adversarial discriminative domain adaptation.
\newblock In {\em CVPR}, volume~1, page~4, 2017.

\bibitem{wang2018low}
Y.-X. Wang, R.~Girshick, M.~Hebert, and B.~Hariharan.
\newblock Low-shot learning from imaginary data.

\bibitem{cub}
P.~Welinder, S.~Branson, T.~Mita, C.~Wah, F.~Schroff, S.~Belongie, and
  P.~Perona.
\newblock Caltech-ucsd birds 200.
\newblock 2010.

\bibitem{latem}
Y.~Xian, Z.~Akata, G.~Sharma, Q.~Nguyen, M.~Hein, and B.~Schiele.
\newblock Latent embeddings for zero-shot classification.
\newblock In {\em CVPR}, pages 69--77, 2016.

\bibitem{goodbadugly}
Y.~Xian, C.~H. Lampert, B.~Schiele, and Z.~Akata.
\newblock Zero-shot learning-a comprehensive evaluation of the good, the bad
  and the ugly.
\newblock {\em TPAMI}, 2018.

\bibitem{featgen}
Y.~Xian, T.~Lorenz, B.~Schiele, and Z.~Akata.
\newblock Feature generating networks for zero-shot learning.
\newblock In {\em CVPR}, 2018.

\bibitem{zhu2017unpaired}
J.-Y. Zhu, T.~Park, P.~Isola, and A.~A. Efros.
\newblock Unpaired image-to-image translation using cycle-consistent
  adversarial networks.
\newblock In {\em CVPR}, pages 2223--2232, 2017.

\end{thebibliography}
}

\end{document}